\documentclass{article}
\usepackage{spconf,amsmath,graphicx}
\usepackage{float}

\title{A Saak Transform Approach to Efficient, Scalable and Robust \\
Handwritten Digits Recognition}
\name{Yueru Chen, Zhuwei Xu, Shanshan Cai, Yujian Lang and C.-C. Jay Kuo}
\address{Ming Hsieh Department of Electrical Engineering\\
University of Southern California, Los Angeles, California, USA}
\begin{document}
\ninept

\maketitle

\begin{abstract}
An efficient, scalable and robust approach to the handwritten digits
recognition problem based on the Saak transform is proposed in this
work.  First, multi-stage Saak transforms are used to extract a family
of joint spatial-spectral representations of input images. Then, the
Saak coefficients are used as features and fed into the SVM classifier
for the classification task. In order to control the size of Saak
coefficients, we adopt a lossy Saak transform that uses the principal
component analysis (PCA) to select a smaller set of transform kernels.
The handwritten digits recognition problem is well solved by the
convolutional neural network (CNN) such as the LeNet-5. We conduct
a comparative study on the performance of the LeNet-5 and the
Saak-transform-based solutions in terms of scalability and robustness 
as well as the efficiency of lossless and lossy Saak transforms under
a comparable accuracy level.
\end{abstract}

\section{Introduction}\label{sec:intro}

Handwritten digits recognition is one of the important tasks in pattern
recognition. It has numerous applications such as mail sorting, bank
check processing, etc.  It is also a challenging task due to a wide
range of intra-class and inter-class variabilities arising from various
writing styles and different handwriting quality. A large number of
methods have been proposed to solve this problem. Methods based on the
convolutional neural network (CNN) offer the state-of-the-art
performance in handwritten digits recognition nowadays. Besides
handwritten digits recognition, we have seen a resurgence of the CNN
methodology \cite{krizhevsky2012imagenet, lecun2015lenet, Nature2015,
wan2013regularization} and its applications to many computer vision
problems in recent years. 

Generally speaking, a CNN architecture consists of a few convolutional
layers followed by several fully connected (FC) layers.  The
convolutional layer is composed of multiple convolutional operations of
vectors defined on a cuboid, a nonlinear activation function such as the
Rectified Linear Unit (ReLUs) and response pooling. It is used for
feature extraction. The FC layer serves the function of a multilayer
perceptron (MLP) classifier. All CNN parameters (or called filter
weights) are learned by the stochastic gradient descent (SGD) algorithm
through backpropagation. It is well known that CNNs-based methods have
weaknesses in terms of efficiency, scalability and robustness.  First,
the CNN training is computationally intensive.  There are many
hyper-parameters to be finetuned in the backpropagation process for new
datasets and/or different network architectures \cite{he2015convolutional,zhang2016accelerating}. Second, trained CNN
models are not scalable to the change of object class numbers and the
dataset size \cite{mccloskey1989catastrophic,goodfellow2013empirical}. If a network is trained for the certain object classes, we
need to re-train the network again even if the number of object classes
increases or decreases by one. Similarly, if the training dataset size
increases by a small percentage, we cannot predict its impact to the
final performance and need to re-train the network as well. Third, these
CNN models are not robust to small perturbations due to their excess
dependence on the end-to-end optimization methodology \cite{fawzi2017geometric,moosavi2016deepfool,nguyen2015deep}. It is desired to develop an alternative solution to overcome these shortcomings.


Being motivated by CNNs, the Saak ({\bf S}ubspace {\bf a}pproximation
with {\bf a}ugmented {\bf k}ernels) transform was recently proposed by
Kuo and Chen in \cite{kuo2017data}. The Saak transform consists of two
new ingredients on top of traditional CNNs. They are: subspace
approximation and kernel augmentation. The Saak transform allows both
forward and inverse transforms so that it can be used for image analysis
as well as synthesis (or generation). One can derive a family of joint
spatial-spectral representations between two extremes - the full
spatial-domain representation and the full spectral-domain
representation using multi-stage Saak transforms.  Being different with
CNNs, all transform kernels in multi-stage Saak transforms are computed
by one-pass feedforward process. Neither data labels nor backpropagation
is needed for kernel computation. 

In this work, we conduct a comparative study on the performance of the
LeNet-5 and the Saak-transform-based solutions regarding their
accuracy, efficiency, scalability, and robustness.  To achieve higher
efficiency, we adopt the lossy Saak transform using the principal
component analysis (PCA) for subspace approximation so as to control the
number of transform kernels (or the principal component number). As to
scalability, the feature extraction process in the Saak transform
approach is an unsupervised one, and it is not sensitive to the class
number. Furthermore, the lossy Saak transform can alleviate the
influence of small perturbations by focusing on principal components
only. We will conduct an extensive set of experiments on the MNIST
dataset \cite{lecun1998gradient} to demonstrate the above-mentioned
properties.  The rest of this work is organized as follows. The Saak
transform is reviewed and the lossy Saak transform is introduced in Sec.
\ref{sec:saak}.  Comparative study on efficiency, scalability, and
robustness of the CNN approach and the Saak transform approach are
examined in Sec.  \ref{sec:study}.  Finally, concluding remarks are
drawn in Sec.  \ref{sec:conclusion}. 

\section{Saak Transform}\label{sec:saak}

\begin{figure}[htb]
\centering
\centerline{\includegraphics[width=0.7\linewidth]{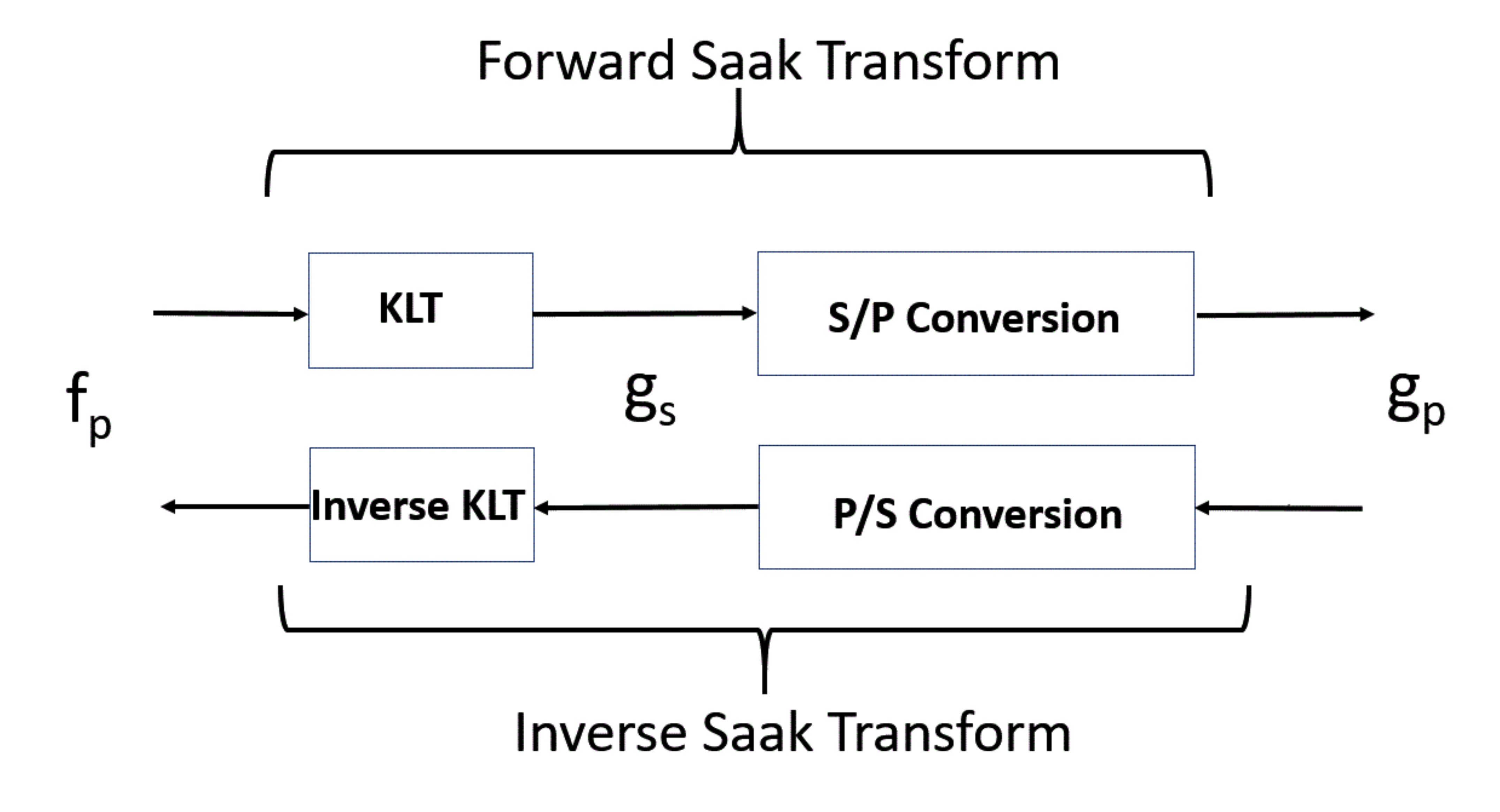}}
\caption{The block diagram of forward and inverse Saak transforms,
where ${\bf f}_p$, ${\bf g}_s$ and ${\bf g}_p$ are the input in
position format, the output in sign format and the output in position
format, respectively.}\label{fig:saak}
\end{figure}

To explain the superior performance of CNNs, Kuo \cite{kuo2016understanding, 
kuo2017cnn} interpreted the CNN architecture as a multilayer RECOS
(REctified-COrrelations on a Sphere) transform.  Furthermore, to define
the inverse RECOS transform that reconstructs an input from its RECOS
transform output as closely as possible, Kuo and Chen \cite{kuo2017data}
proposed the Saak transform recently.  The Saak transform is a mapping
from a real-valued function defined on a 3D cuboid to a 1D rectified
spectral vector. Both forward and inverse Saak transforms can be well
defined.  The 3D cuboid consists of two spatial dimensions and one
spectral dimension.  Typically, the spatial dimensions are set to $2
\times 2$. As to the spectral dimension, it can grow very fast when 
we consider the lossless Saak transform. We will focus on the lossy
Saak transform in this work, which will be elaborated in Sec. 
\ref{sec:study}.

The Saak transforms have several interesting and desired features. First,
it has orthonormal transform kernels that facilitate the computation in
both forward and inverse transforms.  Second, the Saak transform can
eliminate the rectification loss to achieve lossless conversion between
different spatial-spectral representations.  Third, the distance between
two vectors before and after the Saak transform is preserved to a
certain degree.  Fourth, the kernels of the Saak transform are derived
from the second-order statistics of input random vectors. Neither
data labels nor backpropagation is demanded in kernel determination. 

The block diagram of the forward and inverse Saak transforms is given in
Fig.  \ref{fig:saak}. The forward Saak transform consists of three
steps: 1) building the optimal linear subspace approximation with
orthonormal bases using the Karhunen-Lo$\acute{e}$ve transform (KLT)
\cite{stark1986probability}, 2) augmenting each transform kernel with
its negative, 3) applying the rectified linear unit (ReLU) to the
transform output. The second and third steps are equivalent to the
sign-to-position (S/P) format conversion.  The inverse Saak transform is
conducted by performing the position-to-sign (P/S) format conversion
before the inverse KLT.  Generally speaking, the Saak transform is a
process of converting the spatial variation to the spectral variation
and the inverse Saak transform is a process of converting the spectral
variation to the spatial variation. 

\begin{figure}[htb]
\centering
\includegraphics[width=0.9\linewidth]{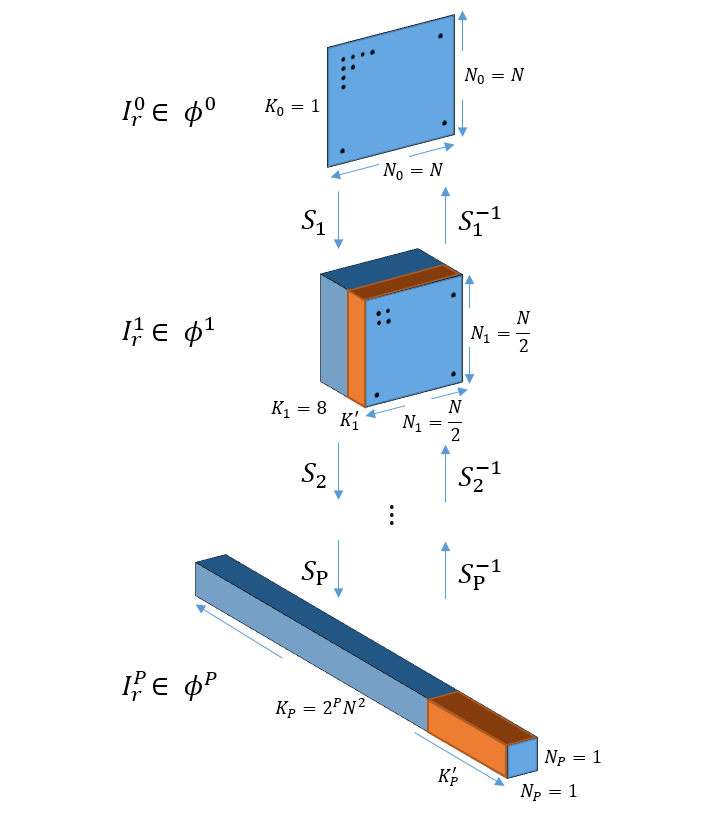} 
\caption{Illustration of the forward and inverse multi-stage Saak
transforms, where $\Phi^p$ is the set of the spatial-spectral
representations in the $p$th stage, and $S_p$ and $S_p^{-1}$ are the
forward and inverse Saak transforms between stages $(p-1)$ and $p$,
respectively.}\label{fig:GC}
\end{figure}

As shown in Fig. \ref{fig:GC}, multi-stage Saak transforms are developed
to transform images of a larger size. To begin with, we decompose an
input into four quadrants recursively to form a quad-tree structure with
its root being the full image and its leaf being a small patch of size
$2 \times 2$. Then, we conduct the Saak transform by merging four child
nodes into one parent node stage by stage and from the leaf to the root.
The whole process terminates when we reach the last stage (or the root
of the tree) that has a spatial dimension of $1 \times 1$. The signed
KLT coefficients in each stage are called the Saak coefficients that can
serve as discriminant features of the input image. Multi-stage Saak
transforms provide a family of spatial-spectral representations.  They
are powerful representations to be used in many applications such as
handwritten digits recognition, object classification, and image
processing. 

Only the lossless Saak transform was examined in \cite{kuo2017data}.
For the lossless multi-stage Saak transforms, the spatial-spectral
dimension in Stage $p$ is $2 \times 2 \times K_p$, where $K_p$ can be
recursively computed as
\begin{equation}\label{eqn:LC}
K_p= 2 \times 4 \times K_{(p-1)}, \quad K_0=1, \quad p=1,2, \cdots.
\end{equation}
The right-hand-side (RHS) consists of three terms. The first term is due
to the S/P conversion. The second and third terms are from that the
degree of freedom of the input cuboid and the output vector should be
preserved through the KLT.  As a consequence of Eq. (\ref{eqn:LC}), we
have $K_p=8^p$. To avoid this exponential growth in terms of the stage
number $p$, we can leverage the energy compaction property of the KLT
and adopt the lossy Saak transform by replacing the KLT with the
truncated KLT (or PCA).  

Many useful properties are preserved in the lossy Saak transform such as
the orthonormality of transform kernels and its capability to provide a
family of spatial-spectral representations. However, there is a loss in
reconstructing the original input. This is the reason that it is called
the ``lossy" transform.  Since we deal with the classification problem,
the exact reconstruction is of little concern here. To reduce the number
of transform kernels, we select the leading $K^{'}_p$ components that
have the largest eigenvalues from all $K_p$ components in stage $p$.
Empirically, we have $K^{'}_p << K_p$. This will be reported in Sec.
\ref{sec:study}. 

\begin{figure}[htb]
\centering
\includegraphics[width=0.8\linewidth]{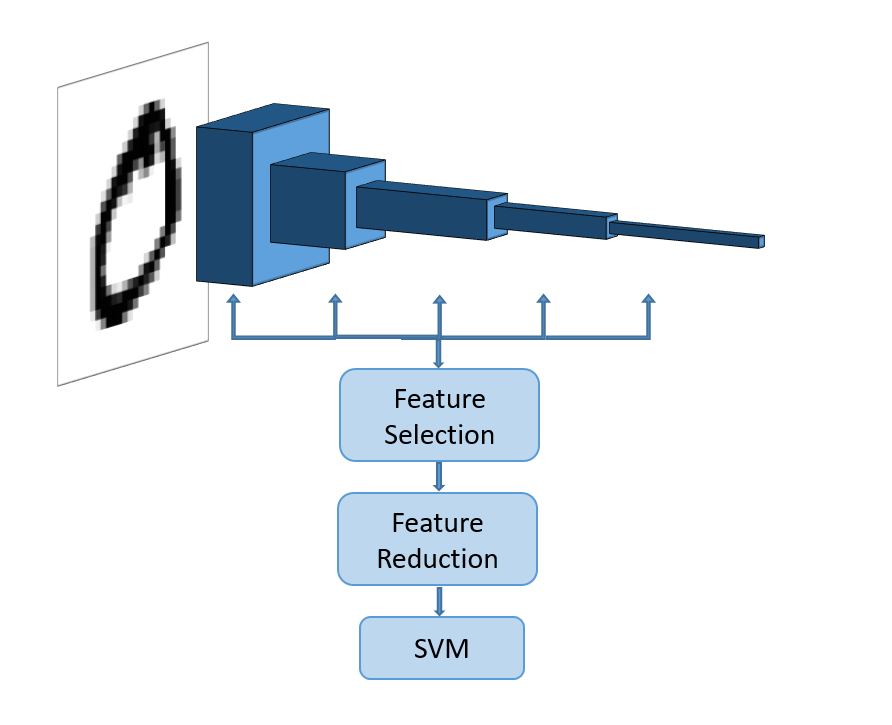} 
\caption{Illustration of the proposed Saak transform approach for
pattern recognition. First, a subset of Saak coefficients is selected
from each stage. Next, the feature dimension is further reduced.
Finally, the reduced feature vector is sent to an SVM
classifier.}\label{fig:CLF}
\end{figure}

\begin{table}[htb]
\normalsize
\centering
\caption{The classification accuracy (\%) of the Saak transform approach
for the MNIST dataset, where the first column indicates the kernel
numbers used from stages 1-5 in the feature selection module while the
second to the fifth columns indicate dimensions of the reduced feature
dimension. The cutoff energy thresholds for the 2nd to 5th rows are 1\%,
3\%, 5\% and 7\% of the total energy, respectively.}\label{table:accuracy_1}
\begin{tabular}{|c|c|c|c|c|}\hline
\#Kernels for each stage     & 32 & 64 & 128 & 256     \\ \hline 
All kernels  & 98.19 & 98.58 & 98.53 & 98.14    \\  \hline
(4, 11, 16, 20, 17)   & 98.24 & 98.54 & 98.33 & 97.84   \\  \hline
(4, 5, 8, 7, 9)   & 98.30 & 98.54 & 98.26 & 97.68  \\  \hline
(4, 5, 5, 6, 7)   & 98.28 & 98.52 & 98.21 & 97.70  \\  \hline
(4, 4, 4, 5, 5)   & 98.22 & 98.42 & 98.08 & 97.58   \\  \hline
\end{tabular}
\end{table}

\begin{table}[htb]
\normalsize
\centering
\caption{The cosine similarity of transform kernels obtained with
subsets and the whole set of MNIST training data, where the first
column indicates the number of images using for training and the second
to sixth columns indicate the cosine similarity in each stage.} 
\label{table:accuracy_2}
\begin{tabular}{|c|c|c|c|c|c|}\hline
Size   & Stage 1    & Stage 2 & Stage 3 & Stage 4 & Stage 5     \\ \hline 
50000  & 0.9999 & 0.9999 & 0.9999 & 0.9999 & 0.9996    \\  \hline
40000  & 0.9999 & 0.9999 & 0.9999 & 0.9999 & 0.9993    \\  \hline
30000  & 0.9999 & 0.9999 & 0.9999 & 0.9999 & 0.9988    \\  \hline
20000  & 0.9999 & 0.9999 & 0.9999 & 0.9996 & 0.9972   \\  \hline
10000  & 0.9999 & 0.9999 & 0.9997 & 0.9992 & 0.9945     \\  \hline
\end{tabular}
\end{table}

\begin{table}[htb]
\normalsize
\centering
\caption{The effect of the training set sizes on the MNIST dataset
classification accuracy where the first row indicates the number of
images used in transform kernel training and the second row indicates
the classification accuracy (\%).}\label{table:accuracy_3}
\begin{tabular}{|c|c|c|c|c|c|c|}\hline
Size      & 60000 & 5000  & 40000 & 30000 & 20000 & 10000  \\ \hline 
Accuracy  & 98.54 & 98.53 & 98.53 & 98.53 & 98.52 & 98.52  \\  \hline
\end{tabular}
\end{table}

\begin{table}[htbp]
\normalsize
\centering
\caption{The cosine similarity of transform kernels using fewer class
numbers and the whole ten classes, where the first column indicates the
object class number in training, and the second to sixth columns
indicate the cosine similarity in each stage.}\label{table:accuracy_4}
\begin{tabular}{|c|c|c|c|c|c|}\hline
Class No.   & Stage 1    & Stage 2 & Stage 3 & Stage 4 & Stage 5     \\ \hline 
8  &  0.9998 & 0.9996 & 0.9942 & 0.9940 & 0.9550    \\  \hline
6  &  0.9982 & 0.9983 & 0.9866 & 0.9639 & 0.5586 \\  \hline
4  &  0.9993 &0.9990 & 0.9816 &0.9219 &0.4557  \\  \hline
2  & 0.9390 & 0.9672 & 0.6567 & 0.6694 & 0.3294 \\ \hline
\end{tabular}
\end{table}

\section{Comparative Study}\label{sec:study}

\begin{table*}[h!]
\normalsize
\centering
\caption{The classification accuracy (\%) on noisy images. All methods
are trained on clean images. The first to fourth columns report the
results of adding Salt and pepper noise as increasing noise level. The
fifth to eighth columns display the results of adding Speckle noise,
adding Gaussian noise, replacing background with uniform noise, and
replacing background with texture images, respectively.
\newline}\label{table:accuracy_7}
\begin{tabular}{|*{9}{c|}} \hline
Method   & S\&P 1   & S\&P 2 & S\&P 3 & S\&P 4  & Speckle & Gaussian   & random\_bg & texture\_bg \\ \hline
LeNet-5  & 89.13 & 86.12 & 74.62   & 67.68  & \bf84.10 & 81.75 & 94.11 & 85.59  \\ \hline
AlexNet  & 82.83 & 84.22 & 62.49  & 53.99  & 75.94 & \bf97,63  & \bf98.36 & \bf98.12 \\ \hline
Saak  & \bf95.71 & \bf95.31 & \bf91.16 & \bf87.49  & 83.06 & 94.08  & 94.67 & 87.78 \\ \hline
\end{tabular}
\end{table*}

\begin{figure}[htbp]
\centering
\includegraphics[width=0.8\linewidth]{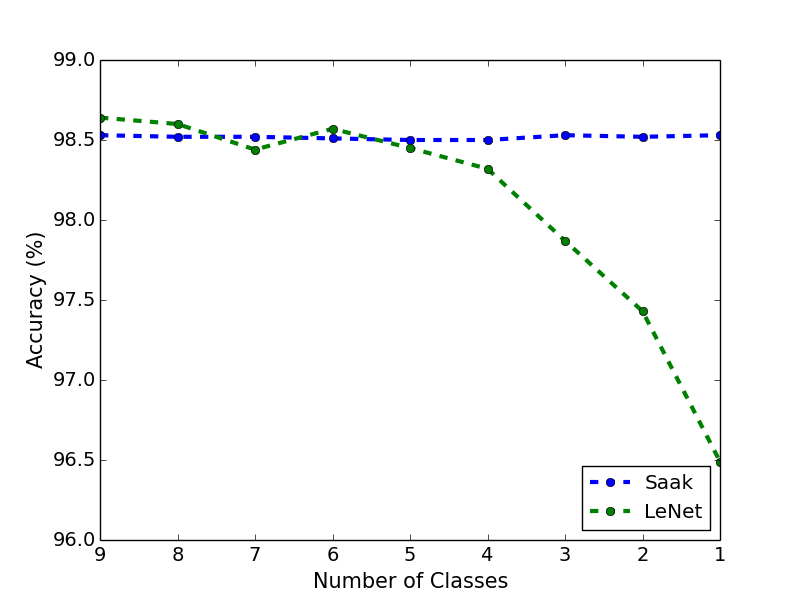} 
\caption{The classification results of using fewer classes in training.
where the blue line indicates the Saak transform approach and the green
line indicates the LeNet-5 method.}\label{fig:ACC}
\end{figure}

We conduct a comparative study on the efficiency of
lossless and lossy Saak transforms under a comparable accuracy level in
Sec. \ref{subsection:efficiency} as well as the performance of the
LeNet-5 and the Saak-transform-based solutions in terms of scalability
and robustness in Secs. \ref{subsection:scalability} and
\ref{subsection:robustness}, respectively. 

\subsection{Efficiency}\label{subsection:efficiency} 

The proposed Saak-transform-based pattern recognition approach is
illustrated in Fig. \ref{fig:CLF}. For an input image of size $32 \times
32$, we conduct five-stage lossy Saak transforms to compute
spatial-spectral representations with spatial resolutions of $16 \times
16 (=256)$, $8 \times 8 (=64)$, $4 \times 4 (=16)$, $2 \times 2 (=4)$
and $1 \times 1 (=1)$.  The feature selection module in the figure
consists of two steps.  For the first step, we select transform kernels
using the largest eigenvalue criterion. This controls the number of Saak
coefficients in each stage effectively while preserving the
discriminative power.  To give an example, if we choose transform
kernels with their energy just greater than 3\% of the total energy of
the PCA, the kernel numbers from Stages 1 to 5 are 4, 5, 8, 7 and 9,
respectively. Then, the number of the Saak features can be computed as
$$
4\times256 + 5 \times 64 + 8 \times 16 + 7 \times 4 + 9 \times 1 
= 1,509,
$$
For the second step, we adopt the F-test statistic (or score) in the
ANalysis Of VAriance (ANOVA) \cite{box1953non} to select features that
have the higher discriminant capability. That is, we order the features
selected in the first step based on their F-test scores from the largest
to the smallest, and select the top 75\%. This will lead to a feature
vector of dimension $1,509 \times 0.75 = 1,132$.  The feature selection
module is followed by the feature reduction module, which is achieved by
the PCA.  We consider four reduced dimension cases of size $32$, $64$,
$128$ and $256$, respectively, as shown in Table \ref{table:accuracy_1}.
Finally, we feed these reduced-dimension feature vectors to the SVM
classifier in the SVM classification module.  We compare different
cutoff energy thresholds in selecting the number of transform kernels in
each stage in Table \ref{table:accuracy_1}.  The first row gives the
lossless Saak transform results and the best one is 98.58\%. We see
little performance degradation by employing the lossy Saak transform,
yet its complexity is significantly reduced. 

\begin{figure}[htbb]
\centering
\includegraphics[width=1.0\linewidth]{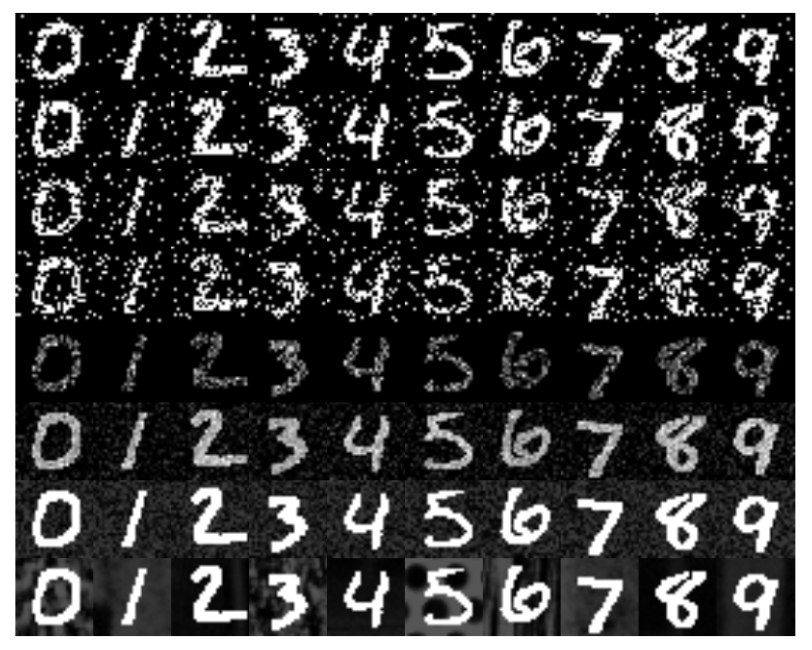} 
\caption{Example of noisy test samples, where the first to fourth rows
are noisy images with added Salt and pepper noise with an increasing
noise level while the fifth to the eighth row are noisy images with
added Speckle noise, Gaussian noise, background replaced by uniform
noise, and background replaced by texture images, respectively.} 
\label{fig:noise}
\end{figure}

\subsection{Scalability} \label{subsection:scalability} 

Here, we adopt multi-stage lossy Saak transforms with the energy
threshold of 3\% and reduce the feature dimension to 64 before applying
the SVM classifier. We compare the average cosine similarity of the Saak
transform kernels using the whole training set and the training subset
in each stage in Table \ref{table:accuracy_2}.  We see from the table
that the transform kernels are very stable, and the cosine similarity
decreases slightly as going to higher stages. Even the lossy Saak
transform is trained using only 10,000 samples, the cosine similarities
between the resulting ones and the one trained using the whole training
set consisting of 60,000 samples are still higher than 0.99. This shows
that the trained kernels are very stable against the training data size.
Next, we report the classification accuracy on the MNIST dataset in
Table \ref{table:accuracy_3} and compare results among training sets of
different sizes. It is interesting to see that we can obtain almost the
same classification accuracy using lossy Saak coefficients based on a
training dataset of a much smaller size. 

The experimental results of scalability against a varying object class
number are shown in Table \ref{table:accuracy_4} and Fig.
\ref{fig:ACC}, where the class is arranged in increasing order. For
example, if there are 4 classes in the training, they are digits 0, 1,
2, 3. We derive transform kernels with a subset of 10 digit classes and
show the averaged cosine similarities in Table \ref{table:accuracy_4}.
As shown in the table, the transform kernels are relatively stable in
the early stages. For example, the averaged cosine similarities of the
first and the second stages are all above 0.9 even we only have one
object class for training.  The cosine similarities for later stages
drop since the Saak coefficients of later stages capture the global view
and they are effected more by decreasing the training classes.  We see
from Fig.  \ref{fig:ACC} that the lossy Saak coefficients learned from
fewer object classes can still handle the classification task of ten
digits. In contrast, we train the convolutional layers of the LeNet-5
under the same above-mentioned condition while it fully connected (FC)
layers are kept the same with ten output classes. Then, the performance
of the LeNet-5 drops quickly when the object class number decreases. The
Saak transform approach is more stable and it recognition accuracy is
about the same even the number of training classes decreases.  In
general, the performance of the Saak transform offers a scalable
solution to the handwritten digits recognition problem because the class
features are obtained from sample class images. The LeNet-5 is more
sensitive since it is built upon the end-to-end optimization framework.
The network has to be re-trained whenever the object class number
changes. 

\subsection{Robustness} \label{subsection:robustness} 

We may encounter undesirable noisy or low quality images in real-world
applications, and need a robust method to handle these situations.  To
simulate the condition, we modify the MNIST dataset using two ways --
adding noise and changing the background of the images as shown in Fig.
\ref{fig:noise}.  To test the robustness of classification methods
fairly, we use only clean images in the training and the unseen noisy
images in the testing. Table \ref{table:accuracy_7} compares the Saak
transform based method, the LeNet-5 and the AlexNet. The LeNet-5
contains 2 CONV layers and 3 FC layers with around 60K parameters. The
AlexNet consists of 5 CONV layers and 3 FC layers with around 60N
parameters. For the Saak transform-based method, we set the energy
threshold to 3\% as described earlier and train the SVM classifier using
32D feature vectors. The results in Table \ref{table:accuracy_7}
indicate that the performance of the lossy Saak transform method is less
affected by noise, especially the Salt and Pepper noise. Although the
Salt and Pepper noise is significantly increased in the column of S\&P
4, our method can still achieve 87.49\% accuracy.  The AlexNet method is
more robust to background change. 

\section{Conclusion and Future Work}\label{sec:conclusion}

A lossy Saak transform based approach was proposed to solve the
handwritten digits recognition problem. This new approach has several
advantages such as higher efficiency than the lossless Saak transform,
scalability against the variation of training data size and object class
numbers and robustness against noisy images. In the near future, we
would like to apply the Saak transform approach to the general object
classification problem with more challenging datasets such as CIFAR-10,
CIFAR-100 and ImageNet. 

\vfill\pagebreak

\newpage

\bibliographystyle{IEEE}
\bibliography{refs}

\end{document}